\documentclass[sigconf]{acmart}
\AtBeginDocument{%
  \providecommand\BibTeX{{%
    \normalfont B\kern-0.5em{\scshape i\kern-0.25em b}\kern-0.8em\TeX}}}

\usepackage{amsmath}

\begin{document}

\title{Highly Generalizable Models for Multilingual Hate Speech Detection}

\author{Neha Deshpande}
\email{neha.n.deshpande@gatech.edu}
\affiliation{%
  \institution{Georgia Institute of Technology}
  \city{Atlanta}
  \state{GA}
  \country{USA}
}

\author{Vidhur Kumar}
\email{vkumar304@gatech.edu}
\affiliation{%
  \institution{Georgia Institute of Technology}
  \city{Atlanta}
  \state{GA}
  \country{USA}
}

\author{Nicholas Farris}
\email{nicholas.farris@gatech.edu}
\affiliation{%
  \institution{Georgia Institute of Technology}
  \city{Atlanta}
  \state{GA}
  \country{USA}
}


\maketitle

\section{Abstract}

Hate speech detection has become an important research topic within the past decade. More private corporations are needing to regulate user generated content on different platforms across the globe. In this paper, we introduce a study of multilingual hate speech classification. We compile a dataset of 11 languages and resolve different taxonomies by analyzing the combined data with binary labels: hate speech or not hate speech. Defining hate speech in a single way across different languages and datasets may erase cultural nuances to the definition, therefore, we utilize language agnostic embeddings provided by LASER and MUSE in order to develop models that can use a generalized definition of hate speech across datasets. Furthermore, we evaluate prior state of the art methodologies for hate speech detection under our expanded dataset. We conduct three types of experiments for a binary hate speech classification task: Multilingual-Train Monolingual-Test,  Monolingual-Train Monolingual-Test and Language-Family-Train Monolingual Test scenarios to see if performance increases for each language due to learning more from other language data. 

\section{Introduction}

Hate speech detection online has increasingly become an important domain of research in the past decade \cite{Fortuna2018}. In the United States, hate speech is protected under the First Amendment, but many other countries including the United Kingdom, Canada, and France have laws prohibiting specific forms of hate speech \cite{rosenfeld2002hate}. With these legal requirements, online platforms must be prepared to internally monitor the content uploaded by users and ensure that they are not enabling the spread of hate speech. The repercussions of unmoderated content could range from large fines to legal action, or even imprisonment. Facebook and Twitter have had to make public statements in response to criticism of their internal monitoring of hate speech to specify what they label as hateful conduct, and how they will be monitoring and regulating this content \footnote{https://help.twitter.com/en/rules-and-policies/hateful-conduct-policy} \footnote{https://transparency.fb.com/policies/community-standards/hate-speech/}. Previous work in machine-learning-based hate speech detection involve developing a discriminator model for a reduced linguistic taxonomy. Previous attempts at multilingual hate speech detection often translate the dataset to English and use models for single language embedding such as LSTM and BERT \cite{Uzan2021}. In this paper, we expand upon prior work in multilingual hate speech detection by making the following contributions:
\begin{enumerate}
    \item Compiling, preprocessing, and quality checking a hate speech dataset consisting of 11 different languages.
    \item Preparing, training, and evaluating deep learning models that effectively recognize hate speech in a multilingual setting and generalize reasonably well to languages not present in the dataset.
    \item Releasing the dataset and trained models to serve as a benchmark for multilingual hate speech detection.
\end{enumerate}

We formalize our experiments by answering the following research questions:
\begin{enumerate}
    \item \textbf{RQ1:} How does hate speech differ from normal speech in a language agnostic setting?
    \item \textbf{RQ2:} How well do the different model architectures perform in hate speech detection when trained on multiple languages?
    \item \textbf{RQ3:} If we trained a model on a particular language family, how well does the model perform on a language that belongs to the same family in comparison to a model trained on the entire set of languages?
\end{enumerate}

\section{Literature Survey/ Baselines}
Corazza et al. \cite{multiling-hate-speech-2} uses datasets for 3 different languages (English, Italian, and German) and trains different models such as LSTMs, GRUs, Bidirectional LSTMs, etc. While this work claims to have a robust neural architecture for hate speech detection across different languages, our work develops models that are \textit{far more generalizable} and trained on a much larger dataset of languages. We also believe that our larger dataset along with novel techniques such as attention-based mechanisms enables us to outperform these models.

Huang et al. \cite{multiling-hate-speech} constructed a multilingual Twitter hate speech corpus from 5 languages that they augmented with demographic information to study the demographic bias in hate speech classification. The demographic labels were inferred using a crowd-sourcing platform. While this study presents an excellent analysis on demographic bias in hate speech detection, it does not leverage the dataset to build a multilingual hate speech classifier, let alone perform any classification tasks.

Aluru et al. \cite{multiling-hate-speech-3} use datasets from 8 different languages and obtain their embeddings using LASER \cite{LASER-code} and MUSE \cite{MUSE-code}. The resultant embeddings are fed in as input to different architectures based on CNNs, GRUs, and variations of Transformer models. While the paper does achieve reasonably good performance across the languages, it falls short on two fronts: 1. The paper claims that the models are generalizable, but these models do not make use of datasets of multiple other languages and lose out on a more fine-tuned set of parameters for generalization. Thus, these models are not expected to perform well outside of the 8 languages, and 2. the study focuses on models that perform well \textit{in low-resource settings} while not paying much attention to how the models perform in an environment not constrained by resources. By combining datasets from 11 different languages we achieve a better performance. 

\section{Dataset Description and Analysis}

\subsection{Source} 
We retrieved our datasets from \textbf{hatespeechdata.com}, a website published by Poletto et al that compiles hate speech data from 11 different languages. Data statistics for the combined datasets for each language is summarized in Table 1. The collection source and taxonomy used for each dataset is described below.

\begin{enumerate}
    \item \textbf{English:} The Ousidhoum et al. \cite{arabic-dataset-2} dataset labels tweets with five attributes (directness, hostility, target, group, and annotator.) The Davidson et al. \cite{english-dataset-1} classifies collected tweets as hate speech or not hate speech using a lexicon directory for increased labeling accuracy. Gilbert et al. \cite{english-dataset-2} uses a text extraction tool to randomly sample posts on Stormfront, a White Supremacist forum, and label them as hate speech or not. The Basile et al. \cite{spanish-dataset-1} dataset categorizes tweets as hate or not hate. Waseem et al. \cite{english-dataset-3} uses the Twitter API to obtain tweets, then classifies them as hate speech or not hate speech. The Founta et al. \cite{english-dataset-4} dataset uses a crowd-sourcing project to compile tweets that are then labeled as hate speech or not hate speech.
    
     \item \textbf{Arabic:} The Arabic Leventine Hate Speech and Abusive Language Dataset \cite{arabic-dataset-1} consists of Syrian/Lebanese political tweets labeled as normal, abusive or hate. Ousidhoum et al. \cite{arabic-dataset-2} dataset of tweets are labeled based on 5 attributes (directness, hostility, target, group, and annotator.)
     
     \item \textbf{German:} Ross et al. \cite{german-dataset-1} compiles a small dataset of tweets regarding refugees in Germany that are classified as hate or normal. Bretschneider et al. \cite{german-dataset-2} crawls 3 German Facebook pages (including comments published) and compiles data annotated as hate or not hate. 
     
    \item \textbf{Indonesian: } Ibrohim et al. \cite{indonesian-dataset-1} collects tweets from Indonesian Twitter for multi-label classification where labels are hate or no-hate, abuse or no-abuse, and attributes of the people targeted including race, gender, religion, etc. The Alfina et al. \cite{indonesian-dataset-2} tweet dataset classifies them as hate speech or not hate speech. The aforementioned datasets were both obtained using a Python Web Scraper and the Twitter API. 
       
    \item \textbf{Italian: } Sanguinetti et al. \cite{italian-dataset-1} dataset of tweets regarding certain minority groups in Italy that are labeled as hate or not hate. Bosco et al. \cite{italian-dataset-2} compiles Facebook posts and tweets about minority groups and immigrants in Italy using the Facebook and Twitter APIs. 
        
    \item \textbf{Portuguese: } Fortuna et al. \cite{portuguese-dataset} compiles hate speech tweets in Portuguese that are labeled at two levels: a simple binary label of hate vs no hate followed by a fine-grained hierarchical multiple label scheme with 81 hate speech categories in total.
        
    \item \textbf{Spanish: } Basile et al. \cite{spanish-dataset-1} includes tweets classified as hate or not hate. Pereira et al. \cite{spanish-dataset-2} compiles tweets that are labeled as hate or not hate. 
         
    \item \textbf{French:} Ousidhoum et al. \cite{arabic-dataset-2} compiles a dataset of tweets that are labeled based on five attributes (directness, hostility, target, group, and annotator.)
         
    \item \textbf{Turkish:} Coltekin et al. \cite{turkish-dataset} randomly samples Turkish blog posts from Twitter and reviews and annotates them as offensive language or not offensive language. Tweets were sampled over a period of 18 months and are regarding various issues currently prominent in Turkish culture.
    
    \item \textbf{Danish:} Sigurbergsson et al. \cite{danish-dataset} introduces \textsc{dkhate}, a dataset of user-generated comments from various social media platforms that are annotated as offensive or not offensive speech. 
    
    \item \textbf{Hindi: } Kumar et al. \cite{hindi-dataset} develops a dataset from Twitter and Facebook that is annotated using binary labels depending on whether or not the example exudes aggression. 
    
\end{enumerate}

\subsection{Data Preprocessing}

\subsubsection{Unifying taxonomy of classification}
One challenge while working with these datasets was different taxonomies of classification. We processed all datasets and re-labeled some to ensure all datasets had consistent binary labels: hate speech or not hate speech. For some datasets, it was relatively straightforward to derive the final binary labels since there was an explicit hate speech label column. For the remaining datasets, we considered all sub-categories of hate speech as hate speech and any categories that didn't fall under the hate speech umbrella as not hate speech. In one special case, for the German dataset by Ross et al. \cite{german-dataset-1} we had to resolve labeling between annotators. For the Ousidham et al. datasets \cite{arabic-dataset-2} in French, English and Arabic which had multi-labeling in terms of sentiment, we considered any data point containing hateful sentiment as hate speech and any containing normal sentiment as not hate speech. Finally, the labels across datasets were mapped to integer labels of 1 for hate speech and 0 for not hate speech.

\subsubsection{Definition of Hate Speech}
Before and during the process of re-labeling the datasets, we uncovered another major challenge: varied definitions of hate speech across datasets. We decided not to define hate speech commonly across datasets, since this may interfere with different nuanced definitions across cultural contexts. Our goal was to create a model that is able to learn a generalized definition of hate speech across all datasets and languages, and hopefully preserve the cultural differences that we are unequipped to identify ourselves.

\subsubsection{General preprocessing}
A few of these datasets: English (Founta et al. \cite{english-dataset-4}), English (Waseem et al. \cite{english-dataset-3}), Italian (Sanguinetti et al. \cite{italian-dataset-1}) consisted only of tweet IDs along with labels for each tweet. We retrieved the corresponding text for each tweet using a Python wrapper for the Twitter API called python-twitter \cite{twitter-api} and appended the column to the original datasets. The Hindi and Arabic datasets were in their original scripts so we transliterated (romanized) them during preprocessing. For Arabic we used Buckwalter transliteration from the lang-trans python package \cite{lang-trans-arabic}, and for Hindi we used the python package indic-transliteration \cite{indic-transliteration} to convert the script to ITRANS romanization scheme. 

General text preprocessing steps included removing carriage return and new line escape characters, removing non-ASCII words, conversion to lowercase, and removing stop words for all supported languages in the nltk python package \cite{nltk}. We also chose to remove emojis because of the differences in connotation, meaning, and usage across different cultures and because we decided to focus on text characteristics to classify data. We combine all datasets within a each language into a single dataset, resulting in 11 datasets for the 11 languages we considered.  We outline the statistics of each language's dataset (number of examples, class imbalance) in the table below:

\renewcommand{\arraystretch}{0.35} 
\begin{table}[h]
    \centering
    \begin{tabular}{p{0.4\linewidth}p{0.2\linewidth}p{0.2\linewidth}}
        \toprule \toprule
        \textbf{Language} & \textbf{\#Examples} & \textbf{\%Hate Speech}\\
        \hline  \\
        English & 65553 & 0.35 \\
        \hline \\
        German & 5568 & 0.26 \\
        \hline \\
        French & 1033 & 0.75 \\
        \hline \\
        Spanish & 10080 & 0.42  \\
        \hline \\
        Italian & 9692 & 0.28  \\
        \hline \\
        Danish & 2619 & 0.12  \\
        \hline \\
        Arabic & 3293 & 0.51  \\
        \hline \\
        Turkish & 27832 & 0.19  \\
        \hline \\
        Portuguese & 4534 & 0.33  \\
        \hline \\
        Hindi & 12000 & 0.36 \\
        \hline \\
        Indonesian & 11104 & 0.41  \\
        \hline \\
        \textbf{Total} & \textbf{157183} \\
        \bottomrule \bottomrule
    \end{tabular}
    \caption{Statistics of each language's dataset. \%Hate Speech refers to the fraction of examples that are hate speech in the dataset, with an ideal value of 0.5.}
    \label{tab:dataset_language_stats}
\end{table}

\section{Experiment Settings and Baselines}
\subsection{The Hate Speech Classification Task}
Our primary task is binary hate speech classification in a single modality, i.e., for only text data. Let $S$ be the raw input text sequence that we want to classify, $Y \in [0, 1]$ be the target variable with $Y = 0$ when $S$ is not hate speech and $Y = 1$ when $S$ is hate speech. We generate a vectorized representation $X = E(S)$ where $E$ is an encoding technique (for example, an embedding generated by LASER \cite{LASER-code} or the output of the encoder in BERT models.) Let $M$ be the model architecture that we train to perform binary hate speech classification. We state the classification problem as follows:
\begin{center}
    \textit{Given $S$, we want our model to output $\hat{Y} = M(X)$ where the predicted label $\hat{Y}$ is ideally the same as $Y$.}
\end{center}

\subsection{Evaluation Metrics}
We chose to use the following evaluation metric to measure model performance: \\
\textbf{Weighted F1 Score:} The F1-score is the harmonic mean of the precision and recall of a classifier. In our case, a lot of the datasets are \textit{class-imbalanced}, which makes the precision metric extremely important. Additionally, the cost of a false negative (predicting something that is hate speech as benign) is high which makes the recall equally important. The F1 score is able to capture both the precision and recall under a \textbf{single, balanced evaluation metric}. We also ensure that the computation of the F1-score is \textit{class-weighted}, i.e., each class is weighted by the fraction of the dataset that it represents.

\subsection{Experiment Parameters}
The system configuration we used for experiments was as follows: 
\begin{enumerate}
    \item \textbf{CPU}: 2 x 2.2GHz vCPU 
    \item \textbf{RAM}: 13 GB DDR4
    \item \textbf{GPU}: 1 x Nvidia Tesla P100
\end{enumerate}
\renewcommand{\arraystretch}{1} 

\subsection{Baselines}
We use the LASER + LR model architecture introduced by Aluru et al. \cite{DE-LIMIT} as our baseline model. The model takes in the raw text sequence as input, generates a 1024 dimensional vector representation of the text from LASER \cite{LASER-code}, and passes the resultant embedding into a Logistic Regression model. We utilize this model in 2 scenarios for our baseline results:
\begin{enumerate}
    \item \textbf{Monolingual-Train Monolingual-Test:} We train the model on a particular language and test it on the same language.
    \item \textbf{Multilingual-Train Monolingual-Test:} We train the model on all the languages and test it on each language separately.
\end{enumerate}
The intuition behind this model is to use LASER as a method of capturing semantic and syntactic relationships of each language and use the representation it outputs to tune the parameters of the logistic regression model.

\section{Proposed Method}
\subsection{Novelty/Improvements}
So far, research in multilingual hate speech detection, suffers from some overarching issues: (1) The \textbf{lack of generalizable models} due to hyper-focused set of languages or insufficient training examples in certain languages, and (2) The \textbf{lack of creativity in leveraging certain properties of each language to their own advantage} (for instance, the similarity of languages within a certain family.) We attempt to overcome these shortcomings by:
\begin{enumerate}
    \item Combining the benefits of an enriched \textit{multilingual} dataset, \textit{cross-lingual} representation methods, and/or complex model architectures to learn parameters using higher-quality representations.
    \item Training complex model architectures on different \textit{language families} to leverage each family's semantic tendencies and achieve relevant parameter learning.
\end{enumerate}

\begin{figure}
    \centering
    \includegraphics[width=\linewidth]{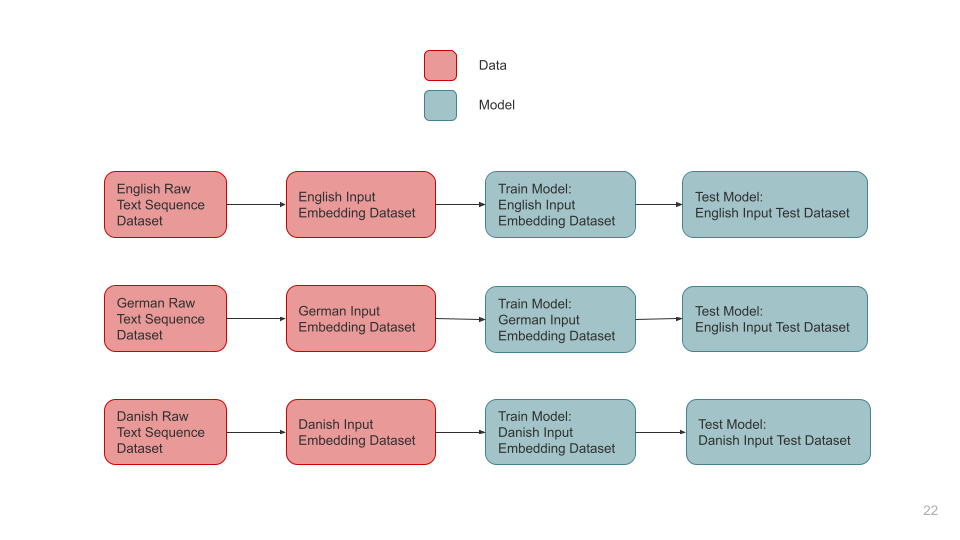} \\
    \vspace{0.5cm}
    \includegraphics[width=\linewidth]{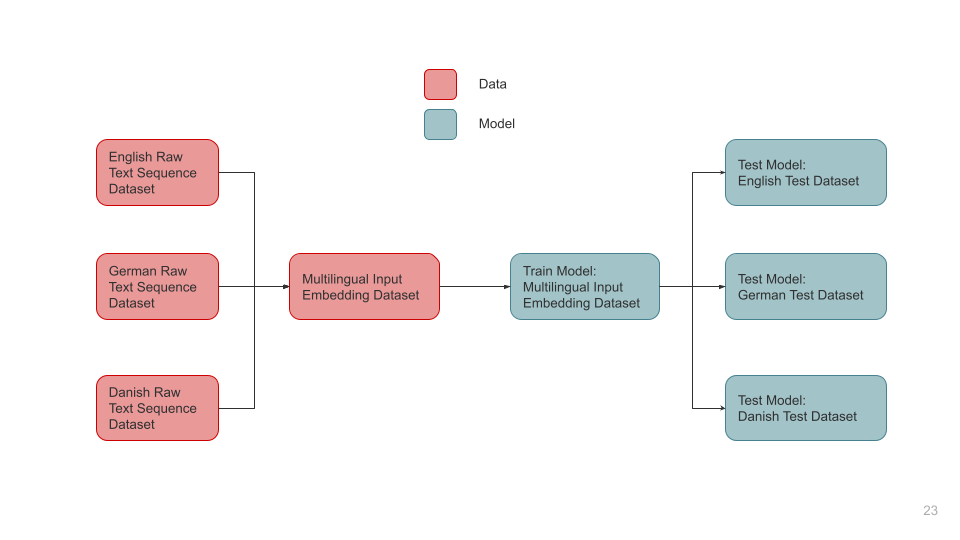}
    \caption{(Top) Monolingual-Train Monolingual-Test visualized for three languages. Here, a separate model is trained for English, German, and Danish and is evaluated on the corresponding language's test dataset. (Bottom) Multilingual-Train Multilingual-Test is visualized for the same three languages. In this case, a single model is trained for all three languages and is evaluated separately on each language's test dataset. The 3 languages are all \textit{Germanic languages} and the model pipeline for our language family experiment is the same as the multilingual scenario.}
    \label{fig:scenarios}
\end{figure}

Formally, we attempt to introduce two novel scenarios as part of our experiments:
\begin{enumerate}
    \item \textbf{Multilingual-Train Monolingual-Test:} We train the model on all available languages and test model performance on each language. 
    \item \textbf{Language Family Train-Test:} We train the model on a particular language family and test model performance on each language in that family.
\end{enumerate}

We train the following model architectures for each scenario in our experiment:
\begin{enumerate}
    \item \textbf{LASER + LR:} We generate sentence-level LASER embeddings and pass them as input to a Logistic Regression model. The model is expected to outperform the SoTA as its trained on our enriched dataset.
    \item \textbf{MUSE + CNN-GRU:} We generate word-level MUSE embeddings for each input word and pass them as input to the CNN-GRU architecture.
    \item \textbf{mBERT:} Multilingual-BERT is a variation of BERT pretrained on 104 languages without their respective language IDs. It can generalize to most of the langauges we consider. We use the raw sentences from each language to fine-tune the mBERT model.
\end{enumerate}

All 3 of the candidate models use a batch size of 16 and are optimized using the AdamW optimizer and Cross Entropy loss. The model-specific hyperparameters are tabulated below:
\begin{table}[h]
    \centering
    \begin{tabular}{cc}
        \toprule \toprule
        \textbf{Hyperparameter} & \textbf{Value} \\
        \hline
        Epochs & 20 \\
        Learning Rate & 0.001 \\
        \bottomrule \bottomrule
    \end{tabular}
    \begin{tabular}{cc}
        \toprule \toprule
        \textbf{Hyperparameter} & \textbf{Value} \\
        \hline
        Epochs & 5 \\
        Learning Rate & 5 x $10^{-5}$ \\
        \# Encoder Layers & 16 \\
        Max. Length & 512 \\
        \bottomrule \bottomrule
    \end{tabular}
    \vspace{0.5cm}
    \begin{tabular}{cc}
        \toprule \toprule
        \textbf{Hyperparameter} & \textbf{Value} \\
        \hline
        Epochs & 20 \\
        Learning Rate & 1 x $10^{-4}$ \\
        \# Conv. Layers & 3 \\
        \# Conv. Filters & 300 \\
        Kernel Sizes & 2x2, 3x3, 4x4 \\
        \bottomrule \bottomrule
    \end{tabular}
    \caption{(Top left, top right, and bottom) Training hyperparameters for LASER + LR, mBERT, and MUSE + CNN-GRU.}
    \label{tab:hyperparams}
\end{table}

\section{Experiments and Results}

\subsection{Monolingual-Train Monolingual-Test}
\textbf{Idea:} Given $n$ languages, we train models $m_1, m_2, \dots, m_n$ for each langauge and test the performance of model $m_i$ on language $l_i$ for $i \in [1, n]$. \\
\begin{table}[h]
    \centering
    \begin{tabular}{ccccc}
        \toprule \toprule
        \textbf{Language} & \textbf{LASER + LR} & \textbf{MUSE + CNN-GRU} & \textbf{mBERT} \\
        \hline
        English & 0.641 & \textbf{0.851} & 0.748 \\
        German & 0.563 & \textbf{0.881} & 0.702 \\
        French & 0.596 & 0.653 & \textbf{0.723} \\
        Spanish & 0.588 & \textbf{0.724} & 0.642 \\
        Italian & 0.686 & \textbf{0.805} & 0.633 \\
        Danish & 0.599 & \textbf{0.853} & 0.808 \\
        Arabic & 0.501 & 0.844 & \textbf{0.898} \\
        Turkish & 0.562 & \textbf{0.790} & 0.729 \\
        Portuguese & 0.645 & 0.838 & \textbf{0.864} \\
        Hindi & 0.697 & \textbf{0.807} & 0.734 \\
        Indonesian & 0.683 & \textbf{0.775} & 0.536 \\
        \bottomrule \bottomrule
    \end{tabular}
    \caption{Weighted F1 scores for all models in the monolingual-train monolingual-test scenario.}
    \label{tab:mono_mono_f1}
\end{table}
\textbf{MUSE + CNN-GRU} significantly outperforms LASER+LR and mBERT for 8/11 languages. mBERT in particular performs worse than both the other models in languages with heavy class imbalances such as Indonesian and Italian. Since mBERT requires a significant amount of data to fine-tune a specific task, it appears from the Monolingual-Train-Tests results that there is not enough data available for each language to properly fine-tune the model. The \textbf{Arabic} dataset has the highest absolute difference in performance, where the LASER+LR F1-score is approximately 60\% lower than the other two models. It is curious that LASER+LR does no better than random guessing for this language while MUSE+CNN-GRU and mBERT are able to achieve a much higher F1 score with only 3300 training examples.

\subsection{Multilingual-Train Monolingual-Test}
\textbf{Idea:} Given $n$ languages, we train a single model $m$ on all $n$ languages and test the model's performance on each langauge $l_i$ for $i \in [1, n]$.

\begin{table}[h]
    \centering
    \begin{tabular}{ccccc}
        \toprule \toprule
        \textbf{Language} & \textbf{LASER + LR} & \textbf{MUSE + CNN-GRU} & \textbf{mBERT} \\
        \hline
        English & 0.672 & 0.854 & \textbf{0.857} \\
        German & 0.577 & \textbf{0.893} & 0.747 \\
        French & 0.590 & 0.721 & \textbf{0.769} \\
        Spanish & 0.573 & 0.669 & \textbf{0.732} \\
        Italian & 0.691 & \textbf{0.824} & 0.661 \\
        Danish & 0.569 & \textbf{0.864} & 0.812 \\
        Arabic & 0.509 & 0.848 & \textbf{0.896} \\
        Turkish & 0.565 & 0.722 & \textbf{0.774} \\
        Portuguese & 0.619 & 0.770 & \textbf{0.887} \\
        Hindi & 0.612 & \textbf{0.798} & 0.739 \\
        Indonesian & 0.687 & \textbf{0.705} & 0.578 \\
        \bottomrule \bottomrule
    \end{tabular}
    \caption{Weighted F1 scores for all models in the multilingual-train monolingual-test scenario.}
    \label{tab:mono_mono_f1}
\end{table} 
\textbf{mBERT} and \textbf{MUSE+CNN-GRU} perform equally well, with mBERT narrowly beating out the latter in 6/11 languages. This is likely because mBERT requires more data to achieve good performance, and with the multilingual dataset, it has access to a richer dataset consisting of 11 languages.  

Compared to the Monolingual-Train, Monolingual-Test scenario, the F1 score on the Monolingual-test scenario with the Multilingual-Train model is better for almost all languages because the model is able to incorporate the semantics of multiple languages to enhance its performance. The F1 score goes down for a few languages which might be due to an undue influence from other languages that are dissimilar and have a large number of examples.

\subsection{Language Families: Germanic and Romance Languages}
\textbf{Idea:} Given $n$ languages, we take a subset of the languages that form a language family $L_F$. We train a model $m_F$ on all languages in $L_F$ and test the model's performance on each language $l \in L_F$.

\begin{table}[h]
    \centering
    \begin{tabular}{cccc}
        \toprule \toprule
        \textbf{Language} & \textbf{Monolingual} & \textbf{Multilingual} & \textbf{Language Family} \\
        \hline
        English & 0.641 & \textbf{0.682} & 0.664 \\
        German & 0.563 & \textbf{0.577} & 0.572 \\
        Danish & 0.569 & \textbf{0.603} & 0.567 \\
    \end{tabular}
    \begin{tabular}{cccc}
        \toprule \toprule
        \textbf{Language} & \textbf{Monolingual} & \textbf{Multilingual} & \textbf{Language Family} \\
        \hline
        English & 0.851 & \textbf{0.854} & 0.842 \\
        German & 0.881 & \textbf{0.893} & 0.811 \\
        Danish & 0.853 & \textbf{0.864} & 0.836 \\
    \end{tabular}
    \begin{tabular}{cccc}
        \toprule \toprule
        \textbf{Language} & \textbf{Monolingual} & \textbf{Multilingual} & \textbf{Language Family} \\
        \hline
        English & 0.748 & \textbf{0.857} & 0.783 \\
        German & 0.702 & \textbf{0.747} & 0.744 \\
        Danish & 0.808 & 0.812 & \textbf{0.815} \\
    \end{tabular}
    \caption{(From top to bottom) Weighted F1 scores for the LASER + LR, MUSE + CNN-GRU, and mBERT models in all three scenarios with a focus on Germanic languages.}
    \label{tab:germanic_f1}
\end{table}

\begin{table}[h]
    \centering
    \begin{tabular}{cccc}
        \toprule \toprule
        \textbf{Language} & \textbf{Monolingual} & \textbf{Multilingual} & \textbf{Language Family} \\
        \hline
        French & \textbf{0.596} & 0.590 & 0.592 \\
        Spanish & 0.588 & 0.573 & \textbf{0.591} \\
        Italian & 0.686 & \textbf{0.691} & 0.688 \\
        Portuguese & 0.645 & 0.619 & \textbf{0.652} \\
    \end{tabular}
    \begin{tabular}{cccc}
        \toprule \toprule
        \textbf{Language} & \textbf{Monolingual} & \textbf{Multilingual} & \textbf{Language Family} \\
        \hline
        French & 0.653 & 0.721 & \textbf{0.746} \\
        Spanish & \textbf{0.724} & 0.669 & 0.694 \\
        Italian & 0.805 & \textbf{0.824} & 0.749 \\
        Portuguese & \textbf{0.838} & 0.770 & 0.780 \\
    \end{tabular}
    \begin{tabular}{cccc}
        \toprule \toprule
        \textbf{Language} & \textbf{Monolingual} & \textbf{Multilingual} & \textbf{Language Family} \\
        \hline
        French & 0.723 & 0.769 & \textbf{0.776} \\
        Spanish & 0.642 & \textbf{0.732} & 0.718 \\
        Italian & 0.633 & 0.661 & \textbf{0.684} \\
        Portuguese & 0.864 & 0.887 & \textbf{0.894} \\
        \bottomrule \bottomrule
    \end{tabular}
    \caption{(From top to bottom) Weighted F1 scores for the LASER + LR, MUSE + CNN-GRU, and mBERT models in all three scenarios with a focus on Romance languages.}
    \label{tab:romance_f1}
\end{table}
From Table 6 and 7, it is evident that for all three model architectures, the Multilingual or Language-Family Train scenarios outperform the Monolingual-Train scenario in most cases. This demonstrates that the richer dataset across all languages allows for better classification performance due to a better generalized idea of hate speech and language semantics. What is interesting to note is: although the Multilingual train scenario tends to outperform the Language Family scenario in most cases, the F1 scores for the Monolingual Test cases are very similar. This indicates that despite having less data, the Language Family model is able to capture more relevant semantic information within a family of languages to achieve a performance close to that of the model trained on all languages. When compared to Germanic languages, the Romance Languages seem to achieve better performance in this scenario. Possible explanations for this could be that the Romance language group has more generalizability between languages compared to the Germanic group. However, we suspect that the lower performance in the Germanic language group can be attributed to the heavy bias of the model towards English with around 80\% of the dataset consisting of English examples. In the future, one could investigate using a sub-sample of the English dataset to identify if this improves performance of the Germanic-language-family-trained model on other languages in the family.

\section{Conclusion}
\subsection{Limitations}
Although we outperform the current SoTA for multilingual hate speech detection models, our work has the following limitations:
\begin{enumerate}
    \item \textbf{Severe Imbalances in the Dataset:} The datasets we use for each language suffers from two major types of imbalances that bias model performance: (1) textbf{Inter-Language Imbalance:} There is a significant difference in the number of examples for each language in our dataset, which biases models towards the semantic tendencies of languages with lots of examples, and (2) \textbf{Class Imbalance:} A number of languages' dataset consists mostly of examples that are \textit{not} hate speech, which may strongly bias neural models.
    \item \textbf{Limited Explainability in Model Performance:} While we have performed a significant number of experiments using various models, our work fails to provide concrete explainability on the model's performance and why one architecture outperforms another for a certain language or scenario. 
\end{enumerate}

\subsection{Extensions}
Multilingual Hate speech detection is an extremely important area of research with lots of potential for improvement. We believe the following to be the most effective extensions to our work:
\begin{enumerate}
    \item \textbf{Data Augmentation to Low-Resource/Imbalanced Languages:} Future work could utilize techniques such as back-translation introduced in Markus et al. \cite{markus-et-al} to augment examples and improve model performance in low-resource/imbalanced languages.
    
    \item \textbf{Novel Candidate Architectures:} Future work could test different architectures such as LASER + CNN-GRU and XLM-RoBERTa to compare performances across models.

    \item \textbf{Zero-Shot/All-but-one Hate Speech Detection:} Another useful extension would be to train models in a multilingual (all-but-one or language families) to detect how well they can generalize to unseen languages.
\end{enumerate}

\section{Contributions}
All of our team members have contributed an equal amount of effort towards making this project successful.
\bibliographystyle{ACM-Reference-Format}
\bibliography{sample-base}


\begin{thebibliography}{33}


\ifx \showCODEN    \undefined \def \showCODEN     #1{\unskip}     \fi
\ifx \showDOI      \undefined \def \showDOI       #1{#1}\fi
\ifx \showISBNx    \undefined \def \showISBNx     #1{\unskip}     \fi
\ifx \showISBNxiii \undefined \def \showISBNxiii  #1{\unskip}     \fi
\ifx \showISSN     \undefined \def \showISSN      #1{\unskip}     \fi
\ifx \showLCCN     \undefined \def \showLCCN      #1{\unskip}     \fi
\ifx \shownote     \undefined \def \shownote      #1{#1}          \fi
\ifx \showarticletitle \undefined \def \showarticletitle #1{#1}   \fi
\ifx \showURL      \undefined \def \showURL       {\relax}        \fi
\providecommand\bibfield[2]{#2}
\providecommand\bibinfo[2]{#2}
\providecommand\natexlab[1]{#1}
\providecommand\showeprint[2][]{arXiv:#2}

\bibitem[\protect\citeauthoryear{??}{ara}{[n.d.]}]%
        {arabic-dataset-1}
 \bibinfo{year}{[n.d.]}\natexlab{}.
\newblock \showarticletitle{L-HSAB: A Levantine Twitter Dataset for Hate Speech
  and Abusive Language}.
\newblock


\bibitem[\protect\citeauthoryear{Aluru, Mathew, Saha, and Mukherjee}{Aluru
  et~al\mbox{.}}{2021}]%
        {DE-LIMIT}
\bibfield{author}{\bibinfo{person}{Sai~Saketh Aluru}, \bibinfo{person}{Binny
  Mathew}, \bibinfo{person}{Punyajoy Saha}, {and} \bibinfo{person}{Animesh
  Mukherjee}.} \bibinfo{year}{2021}\natexlab{}.
\newblock \showarticletitle{A Deep Dive into Multilingual Hate Speech
  Classification}. In \bibinfo{booktitle}{\emph{Machine Learning and Knowledge
  Discovery in Databases. Applied Data Science and Demo Track: European
  Conference, ECML PKDD 2020, Ghent, Belgium, September 14--18, 2020,
  Proceedings, Part V}}. Springer International Publishing,
  \bibinfo{pages}{423--439}.
\newblock


\bibitem[\protect\citeauthoryear{Basile}{Basile}{[n.d.]}]%
        {spanish-dataset-1}
\bibfield{author}{\bibinfo{person}{et.al Basile}.}
  \bibinfo{year}{[n.d.]}\natexlab{}.
\newblock \showarticletitle{SemEval2019-Task5: Multilingual Detection of Hate
  (hatEval)}.
\newblock
\urldef\tempurl%
\url{https://github.com/msang/hateval/tree/master/SemEval2019-Task5}
\showURL{%
\tempurl}


\bibitem[\protect\citeauthoryear{Bayer, Kaufhold, and Reuter}{Bayer
  et~al\mbox{.}}{2021}]%
        {markus-et-al}
\bibfield{author}{\bibinfo{person}{Markus Bayer},
  \bibinfo{person}{Marc{-}Andr{\'{e}} Kaufhold}, {and}
  \bibinfo{person}{Christian Reuter}.} \bibinfo{year}{2021}\natexlab{}.
\newblock \showarticletitle{A Survey on Data Augmentation for Text
  Classification}.
\newblock \bibinfo{journal}{\emph{CoRR}}  \bibinfo{volume}{abs/2107.03158}
  (\bibinfo{year}{2021}).
\newblock
\showeprint[arXiv]{2107.03158}
\urldef\tempurl%
\url{https://arxiv.org/abs/2107.03158}
\showURL{%
\tempurl}


\bibitem[\protect\citeauthoryear{bear}{bear}{[n.d.]}]%
        {twitter-api}
\bibfield{author}{\bibinfo{person}{bear}.} \bibinfo{year}{[n.d.]}\natexlab{}.
\newblock \showarticletitle{A Python wrapper around the Twitter API}.
\newblock
\urldef\tempurl%
\url{https://github.com/bear/python-twitter}
\showURL{%
\tempurl}


\bibitem[\protect\citeauthoryear{Bretschneider}{Bretschneider}{[n.d.]}]%
        {german-dataset-2}
\bibfield{author}{\bibinfo{person}{R Bretschneider, U.;~Peters}.}
  \bibinfo{year}{[n.d.]}\natexlab{}.
\newblock \showarticletitle{Detecting Offensive Statements towards Foreigners
  in Social Media.}
\newblock
\urldef\tempurl%
\url{http://www.ub-web.de/research/}
\showURL{%
\tempurl}


\bibitem[\protect\citeauthoryear{Davidson, Warmsley, Macy, and Weber}{Davidson
  et~al\mbox{.}}{2017}]%
        {english-dataset-1}
\bibfield{author}{\bibinfo{person}{Thomas Davidson}, \bibinfo{person}{Dana
  Warmsley}, \bibinfo{person}{Michael Macy}, {and} \bibinfo{person}{Ingmar
  Weber}.} \bibinfo{year}{2017}\natexlab{}.
\newblock \showarticletitle{Automated Hate Speech Detection and the Problem of
  Offensive Language}. In \bibinfo{booktitle}{\emph{Proceedings of the 11th
  International AAAI Conference on Web and Social Media}} (Montreal, Canada)
  \emph{(\bibinfo{series}{ICWSM '17})}. \bibinfo{pages}{512--515}.
\newblock


\bibitem[\protect\citeauthoryear{de~Gibert, Perez, Garc{\'\i}a-Pablos, and
  Cuadros}{de~Gibert et~al\mbox{.}}{2018}]%
        {english-dataset-2}
\bibfield{author}{\bibinfo{person}{Ona de Gibert}, \bibinfo{person}{Naiara
  Perez}, \bibinfo{person}{Aitor Garc{\'\i}a-Pablos}, {and}
  \bibinfo{person}{Montse Cuadros}.} \bibinfo{year}{2018}\natexlab{}.
\newblock \showarticletitle{{Hate Speech Dataset from a White Supremacy
  Forum}}. In \bibinfo{booktitle}{\emph{Proceedings of the 2nd Workshop on
  Abusive Language Online ({ALW}2)}}. \bibinfo{publisher}{Association for
  Computational Linguistics}, \bibinfo{address}{Brussels, Belgium},
  \bibinfo{pages}{11--20}.
\newblock
\urldef\tempurl%
\url{https://doi.org/10.18653/v1/W18-5102}
\showDOI{\tempurl}


\bibitem[\protect\citeauthoryear{et.al}{et.al}{[n.d.]a}]%
        {indonesian-dataset-2}
\bibfield{author}{\bibinfo{person}{Alfina et.al}.}
  \bibinfo{year}{[n.d.]}\natexlab{a}.
\newblock \showarticletitle{Hate Speech Detection in Indonesian}.
\newblock


\bibitem[\protect\citeauthoryear{et.al}{et.al}{[n.d.]b}]%
        {italian-dataset-2}
\bibfield{author}{\bibinfo{person}{Bosco et.al}.}
  \bibinfo{year}{[n.d.]}\natexlab{b}.
\newblock \showarticletitle{Hate Speech Detection Tasks}.
\newblock
\urldef\tempurl%
\url{https://github.com/msang/haspeede}
\showURL{%
\tempurl}


\bibitem[\protect\citeauthoryear{et.al}{et.al}{[n.d.]c}]%
        {multiling-hate-speech-2}
\bibfield{author}{\bibinfo{person}{Corazza et.al}.}
  \bibinfo{year}{[n.d.]}\natexlab{c}.
\newblock \showarticletitle{A Multilingual Evaluation for Online Hate Speech
  Detection}.
\newblock


\bibitem[\protect\citeauthoryear{et.al}{et.al}{[n.d.]d}]%
        {multiling-hate-speech}
\bibfield{author}{\bibinfo{person}{Huang et.al}.}
  \bibinfo{year}{[n.d.]}\natexlab{d}.
\newblock \showarticletitle{Multilingual Twitter Corpus and Baselines for
  Evaluating Demographic Bias in Hate Speech Recognition}.
\newblock


\bibitem[\protect\citeauthoryear{et.al}{et.al}{[n.d.]e}]%
        {indonesian-dataset-1}
\bibfield{author}{\bibinfo{person}{Ibrohim et.al}.}
  \bibinfo{year}{[n.d.]}\natexlab{e}.
\newblock \showarticletitle{Multi Label Hate Speech and Abusive Language
  Detection in Indonesian Twitter}.
\newblock
\urldef\tempurl%
\url{https://github.com/okkyibrohim/id-multi-label-hate-speech-and-abusive-language-detection}
\showURL{%
\tempurl}


\bibitem[\protect\citeauthoryear{Fortuna and Nunes}{Fortuna and Nunes}{2019}]%
        {portuguese-dataset}
\bibfield{author}{\bibinfo{person}{João Rocha da Silva Juan Soler-Company
  Leo~Wanner Fortuna, Paula} {and} \bibinfo{person}{Sérgio Nunes}.}
  \bibinfo{year}{2019}\natexlab{}.
\newblock \showarticletitle{A Hierarchically-Labeled Portuguese Hate Speech
  Dataset}. In \bibinfo{booktitle}{\emph{Proceedings of the 3rd Workshop on
  Abusive Language Online (ALW3)}}.
\newblock


\bibitem[\protect\citeauthoryear{Fortuna and Nunes}{Fortuna and Nunes}{2018}]%
        {Fortuna2018}
\bibfield{author}{\bibinfo{person}{Paula Fortuna} {and}
  \bibinfo{person}{S\'{e}rgio Nunes}.} \bibinfo{year}{2018}\natexlab{}.
\newblock \showarticletitle{A Survey on Automatic Detection of Hate Speech in
  Text}.
\newblock \bibinfo{journal}{\emph{ACM Comput. Surv.}} \bibinfo{volume}{51},
  \bibinfo{number}{4}, Article \bibinfo{articleno}{85} (\bibinfo{date}{July}
  \bibinfo{year}{2018}), \bibinfo{numpages}{30}~pages.
\newblock
\showISSN{0360-0300}
\urldef\tempurl%
\url{https://doi.org/10.1145/3232676}
\showDOI{\tempurl}


\bibitem[\protect\citeauthoryear{Founta, Djouvas, Chatzakou, Leontiadis,
  Blackburn, Stringhini, Vakali, Sirivianos, and Kourtellis}{Founta
  et~al\mbox{.}}{2018}]%
        {english-dataset-4}
\bibfield{author}{\bibinfo{person}{Antigoni-Maria Founta},
  \bibinfo{person}{Constantinos Djouvas}, \bibinfo{person}{Despoina Chatzakou},
  \bibinfo{person}{Ilias Leontiadis}, \bibinfo{person}{Jeremy Blackburn},
  \bibinfo{person}{Gianluca Stringhini}, \bibinfo{person}{Athena Vakali},
  \bibinfo{person}{Michael Sirivianos}, {and} \bibinfo{person}{Nicolas
  Kourtellis}.} \bibinfo{year}{2018}\natexlab{}.
\newblock \showarticletitle{Large Scale Crowdsourcing and Characterization of
  Twitter Abusive Behavior}. In \bibinfo{booktitle}{\emph{11th International
  Conference on Web and Social Media, ICWSM 2018}}. AAAI Press.
\newblock


\bibitem[\protect\citeauthoryear{Gudbjartur Ingi~Sigurbergsson}{Gudbjartur
  Ingi~Sigurbergsson}{[n.d.]}]%
        {danish-dataset}
\bibfield{author}{\bibinfo{person}{Leon~Derczynski Gudbjartur
  Ingi~Sigurbergsson}.} \bibinfo{year}{[n.d.]}\natexlab{}.
\newblock \showarticletitle{Offensive Language and Hate Speech Detection for
  Danish}.
\newblock
\urldef\tempurl%
\url{http://www.derczynski.com/papers/danish_hsd.pdf}
\showURL{%
\tempurl}


\bibitem[\protect\citeauthoryear{kariminf}{kariminf}{[n.d.]}]%
        {lang-trans-arabic}
\bibfield{author}{\bibinfo{person}{kariminf}.}
  \bibinfo{year}{[n.d.]}\natexlab{}.
\newblock \showarticletitle{Transliteration Library for Non-Latin Languages
  like Arabic, Japanese}.
\newblock
\urldef\tempurl%
\url{https://github.com/kariminf/lang-trans}
\showURL{%
\tempurl}


\bibitem[\protect\citeauthoryear{Ousidhoum, Lin, Zhang, Song, and
  Yeung}{Ousidhoum et~al\mbox{.}}{2019}]%
        {arabic-dataset-2}
\bibfield{author}{\bibinfo{person}{Nedjma Ousidhoum}, \bibinfo{person}{Zizheng
  Lin}, \bibinfo{person}{Hongming Zhang}, \bibinfo{person}{Yangqiu Song}, {and}
  \bibinfo{person}{Dit-Yan Yeung}.} \bibinfo{year}{2019}\natexlab{}.
\newblock \showarticletitle{Multilingual and Multi-Aspect Hate Speech
  Analysis}. In \bibinfo{booktitle}{\emph{Proceedings of EMNLP}}.
  \bibinfo{publisher}{Association for Computational Linguistics}.
\newblock


\bibitem[\protect\citeauthoryear{Pereira}{Pereira}{[n.d.]}]%
        {spanish-dataset-2}
\bibfield{author}{\bibinfo{person}{et.al Pereira}.}
  \bibinfo{year}{[n.d.]}\natexlab{}.
\newblock \showarticletitle{HaterNet a system for detecting and analyzing hate
  speech in Twitter}.
\newblock
\urldef\tempurl%
\url{https://zenodo.org/record/2592149#.YVSAg55KidY}
\showURL{%
\tempurl}


\bibitem[\protect\citeauthoryear{Poletto}{Poletto}{2021}]%
        {hate-speech-data}
\bibfield{author}{\bibinfo{person}{Basile V. Sanguinetti~M. Poletto, F.}}
  \bibinfo{year}{2021}\natexlab{}.
\newblock \showarticletitle{Resources and benchmark corpora for hate speech
  detection: a systematic review.}. In \bibinfo{booktitle}{\emph{Lang Resources
  and Evaluation 55, 477–523}}. \bibinfo{publisher}{Association for
  Computational Linguistics}.
\newblock
\urldef\tempurl%
\url{https://doi.org/10.1007/s10579-020-09502-8}
\showURL{%
\tempurl}


\bibitem[\protect\citeauthoryear{Research}{Research}{[n.d.]a}]%
        {LASER-code}
\bibfield{author}{\bibinfo{person}{Facebook Research}.}
  \bibinfo{year}{[n.d.]}\natexlab{a}.
\newblock \showarticletitle{LASER: Language-Agnostic SEntence Representations}.
\newblock
\urldef\tempurl%
\url{https://github.com/facebookresearch/LASER}
\showURL{%
\tempurl}


\bibitem[\protect\citeauthoryear{Research}{Research}{[n.d.]b}]%
        {MUSE-code}
\bibfield{author}{\bibinfo{person}{Facebook Research}.}
  \bibinfo{year}{[n.d.]}\natexlab{b}.
\newblock \showarticletitle{MUSE: Multilingual Unsupervised and Supervised
  Embeddings}.
\newblock
\urldef\tempurl%
\url{https://github.com/facebookresearch/MUSE}
\showURL{%
\tempurl}


\bibitem[\protect\citeauthoryear{Ritesh~Kumar}{Ritesh~Kumar}{[n.d.]}]%
        {hindi-dataset}
\bibfield{author}{\bibinfo{person}{Akshit Bhatia Tushar~Maheshwari
  Ritesh~Kumar, Aishwarya N.~Reganti}.} \bibinfo{year}{[n.d.]}\natexlab{}.
\newblock \showarticletitle{Aggression-annotated Corpus of Hindi-English
  Code-mixed Data}.
\newblock
\urldef\tempurl%
\url{https://arxiv.org/pdf/1803.09402.pdf}
\showURL{%
\tempurl}


\bibitem[\protect\citeauthoryear{Rosenfeld}{Rosenfeld}{2002}]%
        {rosenfeld2002hate}
\bibfield{author}{\bibinfo{person}{Michel Rosenfeld}.}
  \bibinfo{year}{2002}\natexlab{}.
\newblock \showarticletitle{Hate speech in constitutional jurisprudence: a
  comparative analysis}.
\newblock \bibinfo{journal}{\emph{Cardozo L. Rev.}}  \bibinfo{volume}{24}
  (\bibinfo{year}{2002}), \bibinfo{pages}{1523}.
\newblock


\bibitem[\protect\citeauthoryear{Ross, Rist, Carbonell, Cabrera, Kurowsky, and
  Wojatzki}{Ross et~al\mbox{.}}{2016}]%
        {german-dataset-1}
\bibfield{author}{\bibinfo{person}{Bj{\"o}rn Ross}, \bibinfo{person}{Michael
  Rist}, \bibinfo{person}{Guillermo Carbonell}, \bibinfo{person}{Benjamin
  Cabrera}, \bibinfo{person}{Nils Kurowsky}, {and} \bibinfo{person}{Michael
  Wojatzki}.} \bibinfo{year}{2016}\natexlab{}.
\newblock \showarticletitle{{Measuring the Reliability of Hate Speech
  Annotations: The Case of the European Refugee Crisis}}. In
  \bibinfo{booktitle}{\emph{Proceedings of NLP4CMC III: 3rd Workshop on Natural
  Language Processing for Computer-Mediated Communication}}
  \emph{(\bibinfo{series}{Bochumer Linguistische Arbeitsberichte},
  Vol.~\bibinfo{volume}{17})}, \bibfield{editor}{\bibinfo{person}{Michael
  Bei{\ss}wenger}, \bibinfo{person}{Michael Wojatzki}, {and}
  \bibinfo{person}{Torsten Zesch}} (Eds.). \bibinfo{address}{Bochum},
  \bibinfo{pages}{6--9}.
\newblock


\bibitem[\protect\citeauthoryear{Sai Saketh~Aluru and Mukherjee}{Sai
  Saketh~Aluru and Mukherjee}{[n.d.]}]%
        {multiling-hate-speech-3}
\bibfield{author}{\bibinfo{person}{Punyajoy~Saha Sai Saketh~Aluru,
  Binny~Mathew} {and} \bibinfo{person}{Animesh Mukherjee}.}
  \bibinfo{year}{[n.d.]}\natexlab{}.
\newblock \showarticletitle{Deep Learning Models for Multilingual Hate Speech
  Detection}.
\newblock


\bibitem[\protect\citeauthoryear{Sanguinetti, Poletto, Bosco, Patti, and
  Stranisci}{Sanguinetti et~al\mbox{.}}{2018}]%
        {italian-dataset-1}
\bibfield{author}{\bibinfo{person}{Manuela Sanguinetti}, \bibinfo{person}{Fabio
  Poletto}, \bibinfo{person}{Cristina Bosco}, \bibinfo{person}{Viviana Patti},
  {and} \bibinfo{person}{Marco Stranisci}.} \bibinfo{year}{2018}\natexlab{}.
\newblock \showarticletitle{An Italian Twitter Corpus of Hate Speech against
  Immigrants}. In \bibinfo{booktitle}{\emph{Proceedings of the 11th Conference
  on Language Resources and Evaluation (LREC2018), May 2018, Miyazaki, Japan}}.
  \bibinfo{pages}{2798--2895}.
\newblock
\urldef\tempurl%
\url{https://github.com/msang/hate-speech-corpus}
\showURL{%
\tempurl}


\bibitem[\protect\citeauthoryear{tomaarsen}{tomaarsen}{[n.d.]}]%
        {nltk}
\bibfield{author}{\bibinfo{person}{tomaarsen}.}
  \bibinfo{year}{[n.d.]}\natexlab{}.
\newblock \showarticletitle{Natural Language Toolkit (NLTK)}.
\newblock
\urldef\tempurl%
\url{https://github.com/nltk/nltk}
\showURL{%
\tempurl}


\bibitem[\protect\citeauthoryear{Uzan and HaCohen-Kerner}{Uzan and
  HaCohen-Kerner}{2021}]%
        {Uzan2021}
\bibfield{author}{\bibinfo{person}{Moshe Uzan} {and} \bibinfo{person}{Yaakov
  HaCohen-Kerner}.} \bibinfo{year}{2021}\natexlab{}.
\newblock \showarticletitle{Detecting Hate Speech Spreaders on Twitter using
  LSTM and BERT in English and Spanish - Notebook for PAN at CLEF 2021
  Keywords}.
\newblock  (\bibinfo{date}{09} \bibinfo{year}{2021}).
\newblock


\bibitem[\protect\citeauthoryear{vvasuki}{vvasuki}{[n.d.]}]%
        {indic-transliteration}
\bibfield{author}{\bibinfo{person}{vvasuki}.}
  \bibinfo{year}{[n.d.]}\natexlab{}.
\newblock \showarticletitle{Indic Transliteration Tools}.
\newblock
\urldef\tempurl%
\url{https://github.com/indic-transliteration/indic_transliteration_py}
\showURL{%
\tempurl}


\bibitem[\protect\citeauthoryear{Waseem and Hovy}{Waseem and Hovy}{2016}]%
        {english-dataset-3}
\bibfield{author}{\bibinfo{person}{Zeerak Waseem} {and} \bibinfo{person}{Dirk
  Hovy}.} \bibinfo{year}{2016}\natexlab{}.
\newblock \showarticletitle{Hateful Symbols or Hateful People? Predictive
  Features for Hate Speech Detection on Twitter}. In
  \bibinfo{booktitle}{\emph{Proceedings of the NAACL Student Research
  Workshop}}. \bibinfo{publisher}{Association for Computational Linguistics},
  \bibinfo{address}{San Diego, California}, \bibinfo{pages}{88--93}.
\newblock
\urldef\tempurl%
\url{http://www.aclweb.org/anthology/N16-2013}
\showURL{%
\tempurl}


\bibitem[\protect\citeauthoryear{Çağrı Çöltekin}{Çağrı
  Çöltekin}{[n.d.]}]%
        {turkish-dataset}
\bibfield{author}{\bibinfo{person}{Çağrı Çöltekin}.}
  \bibinfo{year}{[n.d.]}\natexlab{}.
\newblock \showarticletitle{A Corpus of Turkish Offensive Language on Social
  Media}.
\newblock
\urldef\tempurl%
\url{https://coltekin.github.io/offensive-turkish/troff.pdf}
\showURL{%
\tempurl}


\end{thebibliography}

\appendix

\end{document}